\def\year{2022}\relax
\documentclass[letterpaper]{article} 
\usepackage{aaai22}  
\usepackage{amssymb}
\usepackage{amsmath}
\usepackage{booktabs}
\usepackage{bbding}
\usepackage{bm}
\usepackage{times}  
\usepackage{helvet}  
\usepackage{courier}  
\usepackage{graphicx} 
\usepackage{subfigure} 
\usepackage[hyphens]{url}  
\usepackage{graphicx} 
\usepackage{natbib}  
\usepackage{caption} 
\DeclareCaptionStyle{ruled}{labelfont=normalfont,labelsep=colon,strut=off} 
\usepackage{algorithm}
\usepackage{algorithmic}

\usepackage{newfloat}
\usepackage{listings}
\floatstyle{ruled}
\newfloat{listing}{tb}{lst}{}
\floatname{listing}{Listing}
\title{Knowledge Prompting for Few-shot Action Recognition}
\author{
    Yuheng Shi,
    Xinxiao Wu,
    Hanxi Lin
}
\affiliations{
    Beijing Laboratory of Intelligent Information Technology \\
    School of Computer Science, Beijing Institute of Technology, Beijing, China\\
    \{shiyuheng,wuxinxiao,hxlin\}@bit.edu.cn

}

\usepackage{bibentry}

\begin{document}

\maketitle

\begin{abstract}
Few-shot action recognition in videos is challenging for its lack of supervision and difficulty in generalizing to unseen actions. 
To address this task, we propose a simple yet effective method, called knowledge prompting, which leverages commonsense knowledge of actions from external resources to prompt a powerful pre-trained vision-language model for few-shot classification. 
We first collect large-scale language descriptions of actions, defined as text proposals, to build an action knowledge base. The collection of text proposals is done by filling in handcraft sentence templates with external action-related corpus or by extracting action-related phrases from captions of Web instruction videos.
Then we feed these text proposals into the pre-trained vision-language model along with video frames to generate matching scores of the proposals  to each frame, and the scores can be treated as action semantics with strong generalization. Finally, we design a lightweight temporal modeling network to capture the temporal evolution of action semantics for classification. 
Extensive experiments on six benchmark datasets demonstrate that our method generally achieves the state-of-the-art performance while reducing the training overhead to 1\textperthousand\  of existing methods.
\end{abstract}

\section{Introduction}


Few-shot action recognition in videos aims to classify new action classes by using very few training samples. To address this task, the majority of existing works~\cite{bishay2019tarn,Cao2020,Zhu2018,Zhang2020,perrett2021temporal,ben2021taen} formulate the few-shot recognition problem in a meta-learning paradigm, where meta-metrics of  similarity between actions are first trained in the training phase, and then applied for the nearest neighbor voting to make predictions in the test phase.
Although these methods have achieved promissing performance on many datasets like Kinetics~\cite{carreira2017quo}, they still suffer from the very scarce labeled training data that limits their ability to generalize to seldom seen or even unseen action classes. 

In this paper, we provide a fascinating insight that efficiently adapts one well pre-trained vision-language model  to realize the few-shot action recognition task with minimal training. The motivation behind this insight is the superior ability of generalization of the pre-trained vision-language model to novel tasks after it has seen tremendous image-text or vide-text pairs in pre-training. Therefore, we propose a simple yet effective method, called knowledge prompting, which explores commonsense knowledge of actions from external resources to prompt the pre-trained vision-language model well for few-shot recognition. In this work, we employ  CLIP~\cite{radford2021learning} as the pre-trained vision-language model.

To be more specific, we first build an action knowledge base by collecting large-scale textual descriptions of actions from external resources, named as text proposals, which explicitly describe atomic actions such as fine-grained movements of body parts. 
To ensure the knowledge base to cover as many action descriptions as possible, we propose two strategies to generate abundant and various text proposals.
One strategy is to first create the sentence template of ``subject-verb-object”, and then  fill the template with various action-related words  from the external corpus. The corpus consists of the body motion concepts from the PaStaNet dataset~\cite{li2020pastanet} and the object categories from the Visual Genome dataset~\cite{krishna2017visual}. The text proposals generated in this way mainly describe basic actions, and are used as a body of our knowledge base. 
The other strategy is to design a text proposal network to extract action-related phrases from the captions of Web instruction videos, which generates more descriptions of daily actions, thus enriching the text proposals in the knowledge base. 

We take the text proposals and the  video frames as  inputs of the text encoder and the image encoder of CLIP, respectively. For each frame, the output  matching scores  measure how similar the text proposals are to the visual content, and can be treated as potentially valuable representations of action semantics with strong generalization. 
Finally, we design a temporal modeling  network to model the temporal relationships between  the proposal matching scores of different video frames, thereby capturing the evolution of action semantics  along  time for action classification.
Extensive experiments show that the proposed method not only considerably boosts the performance of few-shot action recognition on various datasets, but also greatly reduces the training cost to less than 0.001 times of existing methods.

The main contributions of our work are three-fold:
\begin{itemize}
    \item We propose a knowledge prompting method that steers the pre-trained vision-language model (CLIP) to the few-shot action recognition by leveraging commonsense knowledge from external resources. Our method is simple yet effective, and has strong ability of generalization, without expensive end-to-end training of large-scale backbone.
    \item We propose two strategies of generating abundant and various textual descriptions of actions to build an action knowledge base, and thus well prompt CLIP for learning powerful representations of action semantics. 
    \item We design a lightweight temporal modeling network to model the temporal evolution of action semantics, which further boosts the recognition accuracy.
\end{itemize}

 \section{Related Work}
 
 \subsection{Pre-Trained Vision-Language Models for Visual Recognition}
Pre-trained vision-language models~\cite{radford2021learning,jia2021scaling} have achieved great success in visual recognition due to the addition of natural language to the supervised learning process.
The core problem of applying these models to downstream tasks is prompt learning \cite{schick2020exploiting}, which is a technique that seeks to exploit the learned knowledge encoded in a pre-trained model without tuning the model itself. 

In the field of object recognition, Zhou \emph{et al.} \cite{zhou2021learning} propose to add learnable contexts to the text input of CLIP 
to learn task-relevant prompts. Cho \emph{et al.} \cite{cho2021unifying} formulate several vision-and-language tasks in a unified generative architecture by fine-tuning one multi-modal pre-trained model using task-specific handcraft prompts. Tsimpoukelli \emph{et al.} \cite{tsimpoukelli2021multimodal} train the vision model to learn to cooperate with the encoded common sense knowledge of the frozen language model to generate open-ended outputs and achieve few-shot learning. 
For action recognition, Wang \emph{et al.} \cite{wang2021actionclip} use handrafted labels as the text input of  CLIP \cite{radford2021learning} and fine-tune the whole pre-trained model. 

Different from the aforementioned methods, our knowledge prompting method takes full advantage of commonsense knowledge of actions and generates large-scale prompts to efficiently adapt the pre-trained CLIP model to few-shot action recognition, which no longer requires fine-tuning any parameter.
 
\subsection{Few-shot Action Recognition}
Many existing methods of few-shot action recognition  concentrate on learning the transferable similarity metrics between actions for the nearest neighbor voting,  due to the lack of training data. 
Some methods~\cite{Zhu2018,bishay2019tarn,Zhang2020,perrett2021temporal} learn fine-grained video representations and use dot product or euclidean distance in the representation space as the similarity metric. Zhu \emph{et al.}~\cite{Zhu2018} propose a compound memory network to memorize key-frame features that are vital for adapting to new tasks. Perrett \emph{et al.} \cite{perrett2021temporal} introduce a Transformer-like architecture to learn an adaptive representation space ( \emph{i.e.} query-specific class prototype) via early fusion between the query video and support videos. 
There is also work on explicitly modeling the intrinsic property of video, such as the temporal order \cite{Cao2020}, to assess the similarity.

More recently, Zhu \emph{et al.} \cite{zhu2021closer} get rid of the meta-learning paradigm and focus on exploiting the powerful pre-trained vision backbones for few-shot action recognition. 
They present a classifier-based baseline method and fine-tune the pre-trained model to learn effective representations. In contrast, our method  neither performs  meta-learning  nor fine-tunes  vision backbones. It prompts the pre-trained vision-language model by leveraging external commonsense knowledge of actions to learn powerful action representations with the supervision of language.

\section{Our Method}

\begin{figure*}[t]
	\centering
	\includegraphics[width=\textwidth]{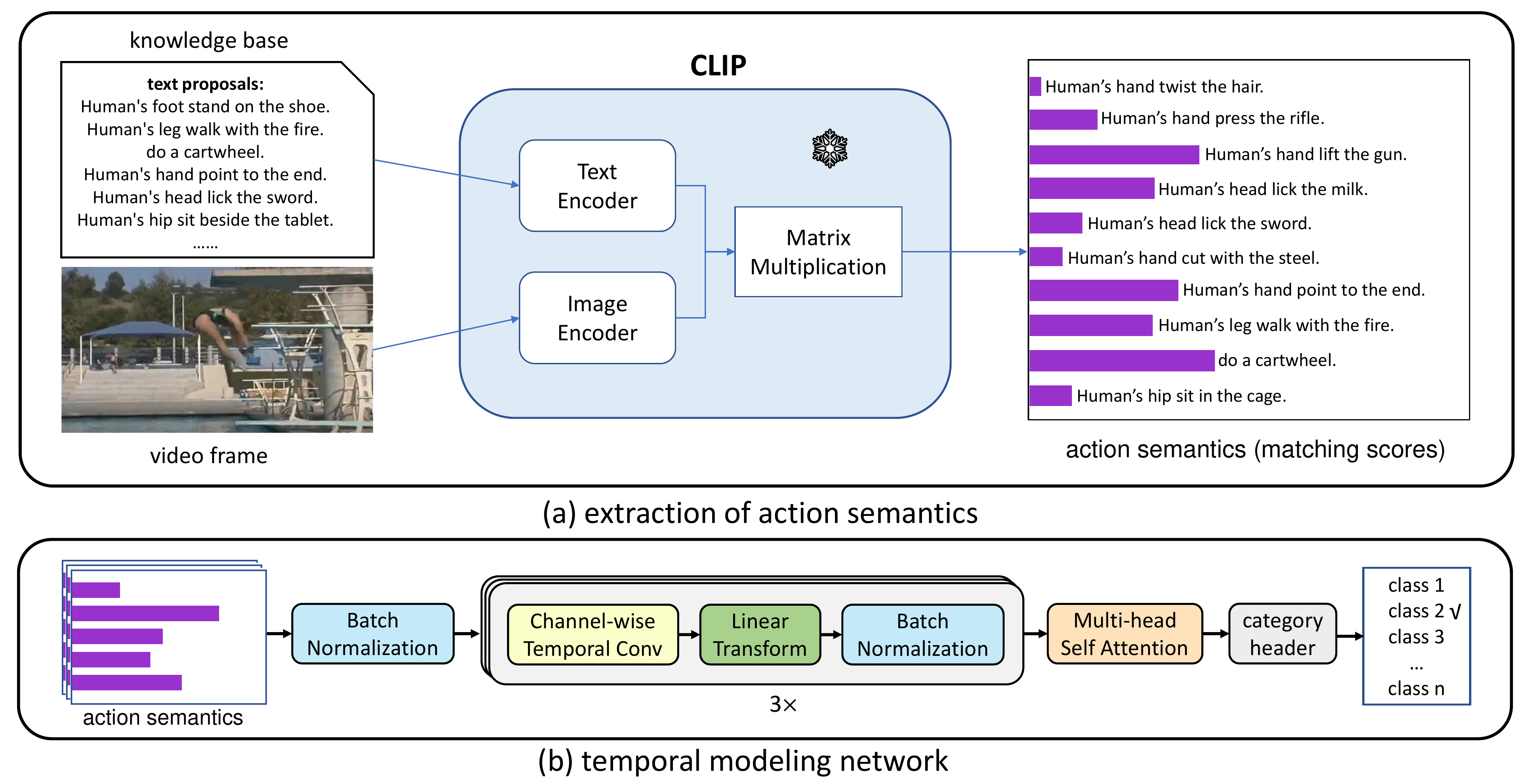}
	\caption{Framework of the proposed method. (a): Extraction of action semantics by CLIP.  (b): Action classification by the temporal modeling network. }
	\label{fig:method_framework}
\end{figure*}

\subsection{Overview}

We propose a knowledge prompting method for few-shot action recognition in videos. It prompts the pre-trained CLIP by using commonsense knowledge from external resources, thereby generalizing well to rare or even unseen actions. The commonsense knowledge is represented by textual descriptions of atomic actions (i.e., text proposals), and an action knowledge base is built by collecting text proposals  from external action-related corpus and video captions.
The core issue of our method lies in how to collect rich and various text proposals for generating semantic representations of actions. To address this issue, we propose two strategies for collecting text proposal: handcraft generation via a sentence template and automatic generation via text proposal network. 

Given an input video, we first take the text proposals as the text input of CLIP, and take video frames as the image input of CLIP. Then, for each video frame, CLIP outputs the similarity matching scores of the text proposals that comprehensively describe the action semantics. Finally, we feed the matching scores of all the video frames into a newly designed temporal modeling network for action classification, by capturing the temporal evolution of action semantics. Figure 1(a) shows the extraction of action semantics by CLIP, and Figure1(b) shows the temporal modeling network for classification.

\subsection{Generation of Text Proposals}

\subsubsection{Handcraft Generation via Sentence Template}
The handcraft generation of text proposals is implemented by first creating the sentence template of ``subject-verb-object" and then filling the template using the action-related words from external corpus. Although currently there is no corpus for directly describing human actions, there are still action-related datasets like  PaStaNet~\cite{li2020pastanet} and Visual Genome~\cite{krishna2017visual}. So we use the body motion concepts from the PaStaNet dataset and the object categories from the Visual Genome dataset as the action-related corpus. 

PaStaNet has a total of 93 states of 10 body parts, such as ``hand, put on" and ``head, kiss", which provides   subjects and verbs  in the sentence  template. 
Visual Genome has dense annotations of  objects and scenes in images, and a total of 5,996 noun words or phrases in the annotations are selected  as objects in the sentence template. In particular,  all transitive verbs or phrases from PaStaNet are paired with nouns or noun phrases from Visual Genome, to  fill in the sentence of ``Human's [body part] [state] the [object]". For example, the body part state ``foot, run to" and the noun ``bed" are used to generate the text proposal ``Human's foot run to the bed".

In this way, we have  380,000 initial text proposals. However, they can not be directly fed into CLIP, since  some linguistically unreasonable proposals will hurt the performance and the  high dimension of matching score vector will make the computation very expensive. So we use a pre-trained mask-based language model, BERT~\cite{devlin2018bert}, to filter the text proposals.
In particular, we mask  the object part (nouns) in the text proposals, and use  BERT to calculate the probabilities of the masked nouns according the subject and verb. If the probability is lower than a threshold $\lambda$ (the value of $\lambda$ will be analyzed in the experiments), 
the corresponding proposal will be discarded. 
For example, for the masked proposal ``Human's foot stand on the [MASK]", we tend to discard the nouns ``code" and ``lincense" with lower probabilities and adopt the nouns   ``bed" and  ``wood"  with higher probabilities.
Finally, we collect more than 50,000 text proposals as the main body of the knowledge base. Figure 2 (a) illustrates the process of the handcraft generative via sentence template. 
\subsubsection{Automatic Generation via Text Proposal Network}

To generate more diverse text proposals to further improve the scalability of the knowledge base, we propose a text proposal network (TPN) that automatically extracts text proposals of daily actions from the action-related captions of Web instruction videos.
It  takes  video captions as input, and outputs action description phrases  as the text proposals. TPN consists of a BERT model to extract token feature of the input sentence, and a classifier to judge whether or not a token belongs to the output text proposal.



To collect the captions of instruction videos from Web, we use query keywords like ``how to",``tutorial" and ``teach" to search action-related instruction videos such as diving and gymnastics tutorial videos from Youtube, and crawl the corresponding captions that have abundant action descriptions. 
To train TPN, we sample 10 captions with about 50,000 words, and annotate the words using the BIO format annotation method~\cite{ramshaw1999text}. Specifically, for each action description (i.e., phrases or sentences) in the captions, ``B" is used to label the first word, and ``I" is used to label the other words in inside. For other descriptions that do not describe actions, ``O" is used to label them. Moreover, we define   two types of action descriptions:  the instance-level description to describe the whole-body movements like ``do a cartwheel", and the part-level description to describe the body-part movements like ``brings his feet together". 

By applying the trained TPN to the instruction video captions, we generate  about 4,000  text proposals with 2,000 proposals for instance-level actions and 2,000 for part-level actions, which further enriches the knowledge base. Figure 2 (b) illustrates the process of the automatic generation via text proposal network. Figure 2 (c) shows several examples of the generated proposals by the two strategies. 

\begin{figure*}[t]
	\centering
	\includegraphics[width=\textwidth]{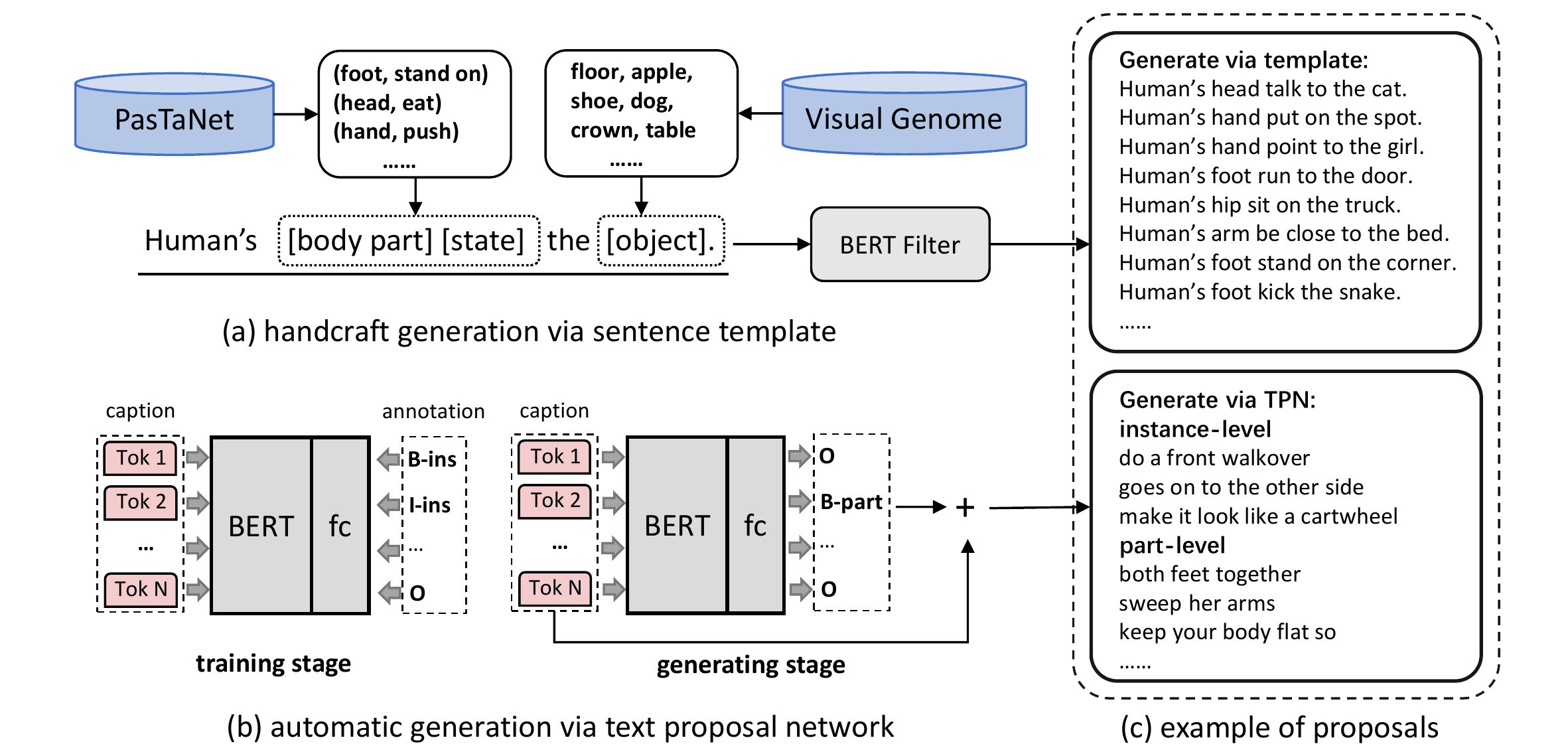}
	\caption{Overview of text proposal generation. (a): Handcraft generation by filling the template using action-related words from PaStaNet and Visual Genome, and filtering out candidates with low probability via BERT. (b): Automatic generation by text proposal network that extract action-related parts from the captions of Web instruction videos. (c): Examples of text proposals generated by the two strategies. }
	\label{fig:knowledge_base}
\end{figure*}

\subsection{Temporal Modeling of Action Semantics }


The generated text proposals are taken as input of the text encoder in CLIP, and the video frames are fed into the image encoder in CLIP. Owing to the great potential of CLIP in bridging the two modalities of vision and language, the output similarity matching scores of the text proposal actually represent the action semantics of each video frame by leveraging commonsense knowledge of actions. 

To capture the temporal relationships between action semantics of different video frames for classification, we propose a temporal modeling network that integrates temporal convolution and multi-head self-attention.

\subsubsection{Extraction of Action Semantics}

Given an input video with $n$ frames $\{f_1, f_2,..., f_n\}$, and a set of $m$ text proposals $\{p_{1}, p_{2} ,...,p_{m}\}$, the pre-trained CLIP model calculates the matching similarities between the  frames and the proposals, denoted as $\bm{S}\in \mathbf{R}^{n\times m}$, where $\bm{S}_{ij}$ represents the matching score between the $i$-th video frame $f_i$ and the $j$-th text proposal $p_{j}$. The higher  $\bm{S}_{ij}$ is, the more relevant  $p_j$ is to $f_i$. The similarity matching scores represent how  the corresponding textual descriptions of actions relate  to the frames, and thus can be treated as  the action semantics of  the frames. Let $\bm{v}_i=[\bm{S}_{i1},\bm{S}_{i2},...,\bm{S}_{im}]$ denote the $i$-th row of $\bm{S}$, and it represents the action semantics of the $i$-th frame. Since the collected text proposals cover rich and various descriptions of atomic actions, the action semantics are more like complete intermediate-level representations of actions with strong generalization. 

It is worth mentioning that the extraction of action semantics does not require training any parameter of CLIP, and we only need to perform the extraction process once for each sample and store the action semantics offline during training. This  differs from other previous methods \cite{wang2021actionclip,perrett2021temporal,Cao2020} that require complete forward and back propagation using the backbone network for each sample in each iteration. Therefore, our method maintains an extremely low computational overhead while achieving state-of-the-art few-shot action recognition performance.

\subsubsection{Temporal Modeling Network}
To capture the temporal contextual relationships between the action semantics to further improve the recognition performance, 
we design a lightweight temporal modeling network (TMN), in which the  action semantics are scaled, combined, time-series modeled, and finally mapped to the action category space.

As illustrated in Figure \ref{fig:method_framework}(b), TMN consists of a batch normalization layer, multiple channel-wise temporal convolution layers, and a multi-head self-attention module. To be more specific, given the sequential action semantics $\{\bm{v}_1,\bm{v}_2,..,\bm{v}_n\}$ as the input of TMN, the batch normalization layer is first employed to  eliminate the distribution bias of  CLIP  for fitting the prior distribution to its training data. Then the multiple channel-wise temporal convolution, linear transformation and batch normalization layers are applied  for the temporal modeling of action semantics. Finally, the multi-head self-attention module is used for global 
temporal modeling the features of all frames and then a linear category header is used for classification.

\begin{table*}[t]
	\small
	\centering 
	\resizebox{1\linewidth}{!}{
		\begin{tabular}{lcccccccc}
			\toprule
			Method                      & Backbone & Kinetics & SS-V2 & HMDB51 & UCF101 & Diving48-V2& FineGym & FLOPs \\ \midrule
			
			CMN \cite{Zhu2018}  &\checkmark    &  78.9 & -  & - & - & - & - &-   \\ 
			CMN-J \cite{zhu2020label} &\checkmark  &  78.9 & -   & - & - &- & -  & -  \\ 
			TARN \citep{bishay2019tarn} &\checkmark  &  78.5 & - & - & - & - & - &-   \\ 
			ARN \cite{Zhang2020}  &\checkmark  &  82.4  &- &  60.6    & 83.1 & - & - &-   \\
			OTAM \cite{Cao2020}  &\checkmark   &  85.8  &52.3 & 72.1 & - & 44.5 & 67.3 &994.3 \\
			TRX \cite{perrett2021temporal}  &\checkmark & 85.9 &64.6  &  75.6    & 96.1  & 62.6 & 72.7 & 1026.7  \\
			STRM \cite{thatipelli2022spatio} &\checkmark   & 91.2  &\bf{70.2} &  81.3    & 98.1 & - & - & 1067.8 \\
			\midrule
			CLIP \cite{radford2021learning} & \XSolidBrush  & 94.2 &26.2 & 31.3 & 93.5 & 22.9 & 20.1 & - \\
			Ours   & \XSolidBrush  &  \bf{94.3} &62.4  & \bf{87.4} & \bf{99.4} &  \bf{82.6}  & \bf{76.8} &\textbf{0.8}  \\
			\bottomrule
		\end{tabular}
	}
	\caption{Comparison results (\%) with the state-of-the-art methods of 5-way 5-shot on the Kinetics, SS-V2, HMDB51, UCF101, Diving48-V2 and FineGym datasets. The "Backbone" column represents whether the method needs to train a backbone in training. The "FLOPs" column represents the amount of floating-point operations for a video sample during training. 
	}
	\label{tab:main_results}
\end{table*}

\section{Experiments}

\subsection{Datasets}

To evaluate our method, we conduct experiments on six action datasets, including Kinetics~\cite{carreira2017quo}, Something Something V2 (SS-V2)~\cite{goyal2017something},  HMDB51~\cite{kuehne2011hmdb}, UCF101~\cite{soomro2012ucf101}, Diving48-V2~\cite{li2018resound}, and FineGym~\cite{shao2020finegym}.
For Kinetics, SS-V2 and UCF101, we adopt the existing splits proposed in~\cite{Zhu2018,Cao2020,Zhang2020}, which consist of 64, 12 and 24 classes in the training, validation and test sets, respectively. 
For HMDB51, we use the split of 31, 10 and 10 classes in the training, validation and test sets, respectively, following~\cite{Zhang2020}.
For Diving48-V2, we use the split of 36, 6 and 6 classes in the training, validation and test sets, respectively.
For FineGym, we use the split of 72, 13 and 14 classes in the training, validation and test sets, respectively.

\subsection{Implement Details}

\begin{table*}[t]
	\small
	\centering 
	\resizebox{0.71\linewidth}{!}{
		\begin{tabular}{lcccccccc}
			\toprule
			Method      & Kinetics & SS-V2 & HMDB51 & UCF101 & Diving48-V2& FineGym \\ \midrule
			w/o knowledge   &  91.7 & 56.0 & 84.7 & 99.0 & 78.8 & 74.0    \\ 
			w/o TPN  & 94.1 &62.2 & 86.8 & \textbf{99.6} & 81.5 & 76.1 \\
			w/o TMN & 93.3 & 49.0 & 85.2 & 99.2 & 63.7 & 69.0    \\
		
		   Ours  &  \bf{94.3} &\textbf{62.4}  & \textbf{87.4} & 99.4 &  \textbf{82.6}  & \bf{76.8}   \\
				\bottomrule
		\end{tabular}
	}
	\caption{Results (\%) of ablation studies on the Kinetics, SS-V2, HMDB51, UCF101, Diving48-V2 and Fine Gym datasets. }
	\label{tab:proposal}
\end{table*}

\begin{table*}[t]
	\small
	\centering 
	\resizebox{0.82\linewidth}{!}{
		\begin{tabular}{lcccccccc}
			\toprule
			Value of $\lambda$ & Proposal Number & Kinetics & SS-V2 & HMDB51 & UCF101 & Diving48-V2 &FineGym \\ \midrule
			$6\times10^{-4}$  & 14388 & 92.1 & 60.0 & 85.1 & 99.0 & 80.1 & 74.8 \\
			$3\times10^{-4}$  & 25172 & 93.0 & 61.3 & 86.1 & 99.1 & 80.8 & 75.0 \\
			$2\times10^{-4}$  & 33763 & 93.5 & 61.6 & 85.3 & 99.2 & 80.7 &\bf{76.1} \\
			$1\times10^{-4}$  & 53133 & \bf{94.1} & \textbf{62.2} & \textbf{86.8} & \textbf{99.6} & \textbf{81.5} & 74.9   \\
			\bottomrule
		\end{tabular}
	}
	\caption{Results (\%) of using the text proposals generated by handcraft sentence template with different $\lambda$ on the Kinetics, SS-V2, HMDB51, UCF101, Diving48-V2 and FineGym datasets. }
	\label{tab:lambda}
\end{table*}

During training, the CLIP model is frozen  and only the temporal modeling network is learned using the training set. During test, the parameters of both CLIP and the temporal modeling network are kept fixed, except for the linear classifier that is fine-tuned for the target classes.
ViT-B/16~\cite{dosovitskiy2020image} is used as the image encoder and the text encoder of CLIP. The pre-processing of image and text feature extraction remains the same as the original CLIP. 
The temporal sparse sampling \cite{wang2018temporal} is adopted to sample video frames  as the input of CLIP, and the number of sampled  frames is set to 16 for each video.

In terms of training parameters, SGD with momentum is used as the optimizer. The learning rate is initially set to 0.001, and is  attenuated by 10 times at 20, 30 and 40 training epochs, respectively. A random dropout layer with a dropout probability of 0.05 is applied after the first batch normalization layer of the temporal modeling network. The momentum coefficient is 0.9, the L2 regularization coefficient is 0.001, and the batch size is 32.

In terms of test settings, Adam is used as the optimizer. The initial learning rate is 0.01 and the exponential decay rate coefficient for moment estimates are set to 0.5 and 0.999.
The training stops after 10 training epochs. 
The L2 regularization coefficient is 0 and the batch size is 16.
The prediction result of a single sample is the average prediction result after randomly sampling the video ten times using the model prediction. The standard 5-way 5-shot evaluation is employed on all datasets and the average accuracy over 500 random test tasks is reported.

\subsection{Experimental Results} 
\subsubsection{Comparison with State-of-the-Art  Methods}
Table~\ref{tab:main_results} shows the comparison results of the standard 5-way 5-shot action recognition task with the state-of-the-art methods on the six action datasets. 
It is interesting to observe that our method generally achieves best results on most datasets with extremely low computational overhead. This benefits from the strong generalization of extracted action semantics  through the collection of abundant text proposals and the powerful vision-language matching ability of CLIP.
Moreover, the freeze of CLIP model parameters and the efficient temporal modeling network design greatly reduce the computational cost of training. 

We can also observe that the proposed method performs not very well on the SS-V2 dataset, probably  due to that most of the actions in SS-V2 are about  fine-grained hand-object interactions such as ``pretending to put something underneath something" and ``moving something across a surface until it falls down", and it is difficult to collect  relevant descriptions of these actions as text proposals.  

\subsubsection{Comparison with Baseline Method}

We also compare our method with a baseline model, called CLIP zero-shot, which directly uses the visual features extracted by the image encoder of CLIP for action recognition and without temporal modeling. 
The  results are shown in the bottom part of Table \ref{tab:main_results}. It is obvious that our method achieves better results on all the datasets, especially on  SS-V2, Diving48-V2 and FineGym, which demonstrates the superiority of our design of extracting action semantics via prompting CLIP using commonsense knowledge and modeling the temporal information of action semantics by TMN. 

\subsection{Ablation Studies}

To study in-depth of different individual components, we introduce several variants of our method for comparison, as follows:
\begin{itemize}
\item 
{\textbf{w/o knowledge}}: We remove the knowledge base  to valuate the contribution of  text proposals. In this case, only the visual features from the image encoder of CLIP are directly fed into the temporal modeling network for classification. 
\item
{\textbf{w/o TPN}}: We remove the  text proposal network to evaluate its effectiveness. In this case, only the text proposals generated by the  sentence template are used.  
\item 
{\textbf{w/o TMN}}: To evaluate the importance of the temporal modeling network, we replace it using  a linear mapping layer along with a batch normalization layer. 
\end{itemize}

\begin{figure*}[t]
	\centering
	\includegraphics[width=\textwidth]{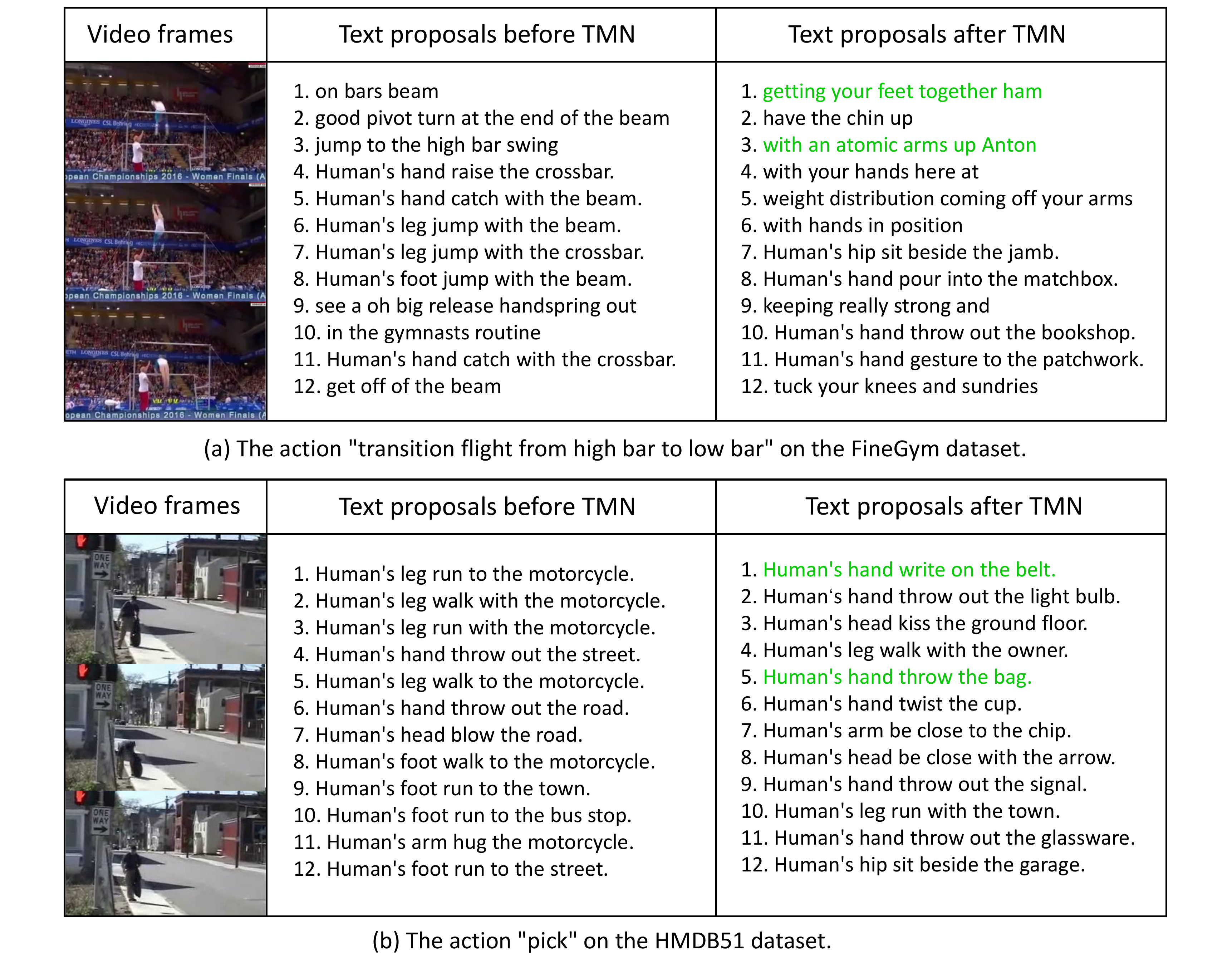}
	\caption{Illustration of several examples of video frames with  most important text proposals before and after the temporal modeling network (TMN). The text proposals are ranked by the matching similarity scores from CLIP before TMN and the gradient sizes corresponding to the action semantic vector after TMN. The proposals metioned in paper have been marked out.}
	\label{fig:interpret_1}
\end{figure*}

The results of ablation studies on the six datasets are shown in Table \ref{tab:proposal}. We have the following observations:
\begin{itemize}
\item
The performance  degrades on all the datasets when removing the text proposals, which validates the benefit of  prompting CLIP using external knowledge to enhancing the generalization ability in few-shot recognition.
\item
When removing the text proposals automatically generated by TPN, the performance also drops on most datasets, which indicates the efficacy of extracting action descriptions from the captions of Web instruction videos on enriching the knowledge base of text proposals. It should be mentioned that the proposals generated by TPN will inevitably contain some noise words since some captions are automatically generated by speech recognition, but experiments show that this doesn't affect their vital role in recognition(illustrated in Figure \ref{fig:interpret_1}).  
\item
By removing the temporal modeling network, our method achieves much worse results, clearly demonstrating that it is essential to capture the temporal relationships between action semantics for action classification.
\end{itemize}

\subsection{Analysis of Threshold $\lambda$}
To analyze the effect of text proposal filtering in handcraft generation, we conduct experiments with different values of the probability threshold $\lambda$.
Table \ref{tab:lambda} show the results of only using the text proposals generated by handcraft sentence template with different $\lambda$, where the larger $\lambda$ represents that more text proposals are filtered out. It can be observed that a smaller $\lambda$ generally achieves fairly better performance, which suggests that the increasing text proposals are helpful to boost the accuracy owing to more supervision from language. For FineGym, the optimal value of $\lambda$ is larger than that for other datasets. The possible reason is that the videos in the FineGym dataset are professional actions of formal gymnastics competitions where the scenes and interactive objects are relatively simple compared to other datasets, so there are less proposals related to frames. In this case, more proposals with low probability bring more distractors, which hurts the performance.

\subsection{Evaluation of Text Proposals}



Figure \ref{fig:interpret_1} illustrates several examples of video frames with a few  most important text proposals before and after TMN, where the importance of text proposals before TMN is determined by the matching similarity scores after CLIP, and the importance of text proposals after TMN is calculated by the gradient sizes corresponding to the action semantic vector in the test phase. 

As shown in Figure \ref{fig:interpret_1}(a), most initially generated text proposals mainly  describe the interaction between ``human" and ``beam" or ``crossbar". That's because the ``beam" or ``crossbar" is really the most obvious object in video frames, which is easily recognized by CLIP without temporal modeling. And it is interesting to observe that some proposals such as ``getting your feet together" and "arms up" become more important after temporal modeling, since they describe the discriminative fine-grained body movements and thus play a vital role in final recognition. 

Figure \ref{fig:interpret_1}(b) further demonstrates the effect of temporal modeling. The video is about a man picking up rubbish from the roadside to his bag. But the shape of the man and the bag is somewhat misleading, making CLIP recognize a motorcycle in the frames and resulting in high scores for the proposals containing ``run" or ``motorcycle". Our model makes the correct classification of ``pick" rather than ``run" in such case, and it can be seen that most important proposals after TMN are closer to dynamically describing the actions like ``Human's hand write on the belt." and ``Human's hand throw the bag". This is due to the reliability of our TMN and the strong ability of generalization of the collected abundant text proposals.

\section{Conclusion}

We have presented a knowledge prompting method that can efficiently adapt a pre-trained vision-language model (CLIP) by leveraging commonsense knowledge from external resources to achieve the few-shot action recognition. To that end, we have proposed two strategies that are able to generate abundant text proposals as the text input of CLIP. A lightweight network is also designed for temporal modeling of action semantics, and  succeeds in boosting the performance. Our method is simple yet effective, with strong ability of generalization and low computational overhead. Extensive experiments on six action datasets demonstrate the effectiveness and superiority of our method on few-shot action recognition. 

But for some specific actions such as fine-grained hand movements in the SS-V2 dataset, the performance of our method is not satisfactory due to  the limited relevant text proposals. So  in future work, we are going to explore more external resources like  textual corpus of actions to further enrich the knowledge base, and meanwhile to introduce uncertainty learning to improve the text proposal prompting.  

\bibliography{aaai22}

\end{document}